\pdfoutput=1
\PassOptionsToPackage{table,xcdraw}{xcolor}
\documentclass[11pt]{article}
\usepackage{authblk}
\usepackage{acl}

\usepackage{times}
\usepackage{latexsym}

\usepackage[T1]{fontenc}
\usepackage[utf8]{inputenc}
\usepackage{multirow}
\usepackage{arydshln}
\usepackage{booktabs}
\usepackage{graphicx}
\usepackage{enumitem}
\usepackage{subfigure}
\usepackage{titlesec}

\usepackage{caption}
\usepackage{bm}

\usepackage[most]{tcolorbox}
\colorlet{Blue}{blue!40!}
\tcbset{on line, 
        boxsep=2pt, left=0pt,right=0pt,top=0pt,bottom=0pt,
        colframe=white,colback=Blue,  
        highlight math style={enhanced}
        }
\newcommand{\paragraphHdNospace}[1] {\noindent\textbf{#1}} % for initial headings (no extra spacing)
\newcommand{\paragraphHd}[1] {\vspace{1.2mm}\noindent\textbf{#1}} % for subsequent headings (small space added)

\usepackage{microtype}

\title{Improving the Generalizability of Depression Detection\\by Leveraging Clinical Questionnaires}

\author[1,2]{\textbf{Thong Nguyen}}
\author[1,2]{\textbf{Andrew Yates}}
\author[3]{\textbf{Ayah Zirikly}}
\author[4]{\textbf{Bart Desmet}}
\author[5]{\textbf{Arman Cohan}}
\affil[1]{University of Amsterdam\\Amsterdam, Netherlands}
\affil[2]{Max Planck Institute for Informatics\\Saarbrücken, Germany}
\affil[3]{Johns Hopkins University\\Baltimore, Maryland}
\affil[4]{National Institutes of Health\\Bethesda, Maryland}
\affil[5]{Allen Institute for AI\\Seattle, WA}
\begin{document}
\maketitle
\begin{abstract}
Automated methods have been widely used to identify and analyze mental health conditions (e.g., depression) from various sources of information, including social media. Yet, deployment of such models in real-world healthcare applications faces challenges including poor out-of-domain generalization and lack of trust in black box models.
%auditability due to large amount of user input. 
%explainability.
%In the healthcare domain, it is especially critical to be able to ground model predictions in intuitive explanations.
In this work, we propose approaches for depression detection that are constrained to different degrees by the presence of symptoms described in PHQ9, a questionnaire used by clinicians in the depression screening process.
In \textit{dataset-transfer} experiments on three social media datasets, we find that grounding the model in PHQ9's symptoms substantially improves its ability to generalize to out-of-distribution data compared to a standard BERT-based approach.
Furthermore, this approach can still perform competitively on in-domain data.
These results and our qualitative analyses suggest that grounding model predictions in clinically-relevant symptoms can improve generalizability while producing a model that is easier to inspect.
%In experiments on three social media datasets, we find approaches grounded in PHQ9's symptoms perform comparably to a standard BERT-based approach.
%that such approaches perform well and even outperform a standard BERT model in \textit{dataset-transfer} evaluations, despite the fact that training and test data 
%In the appendix, we show that our constrained models are superior to GPT-3 in few-shot settings.
%These results suggest that grounding model predictions in PHQ9 symptoms improves generalizability in addition to the auditability benefits.
\end{abstract}

\section{Introduction}
Given the significance of mental health as a public health challenge \citep{51109334-209f-4220-b5de-bfad08259c34}, much work has investigated approaches for detecting mental health conditions using social media text \citep{yates-etal-2017-depression,Coppersmith2018NaturalLP,shing2020prioritization,harrigian-etal-2021-state}.
Such approaches could be used by at-risk users and their clinicians to monitor behavioral changes (e.g., by monitoring changes in the presence of symptoms related to depression as treatment progresses.
These approaches generally rely on datasets consisting of users with self-reported diagnoses (e.g., based on a statement like \textit{``I was just diagnosed with depression''}) for training and evaluation ~\citep[e.g.,][]{yates-etal-2017-depression, cohan2018smhd}.
Despite promising results on these tasks, related work argues that assessing depression and suicidal behavior is difficult in practical settings and even experienced clinicians frequently struggle to correctly interpret signals \cite{Coppersmith2018NaturalLP}.
Furthermore, recent work has found that models trained on particular mental health datasets do not always generalize to others.
%~\citep{harrigian-etal-2020-models}.
\citet{harrigian-etal-2020-models, 10.1145/3290605.3300364} find that systematic, spurious differences between diagnosed and control users can prevent trained models from generalizing to even other, similar social media data.
Similarly, outside the mental health domain, recent work reports that %underspecification hinders
neural models often struggle to generalize to data outside their training distribution~\citep{Geirhos2020ShortcutLI,damour2020underspecification,harrigian-etal-2020-models}. 
% These models contain a huge number of parameters and thus have the freedom to identify features that perform well in the training domain but do not generalize.
% This behavior is related to the generalization difficulties reported by~\citet{harrigian-etal-2020-models}.

%These differences are introduced by the dataset construction process.
%For example, \citet{harrigian-etal-2020-models} found that in some datasets, diagnosed users were more likely to use terms related to gender and sexuality, whereas control users were more likely to use terms associated with sports.

% Outside the task of detecting mental health conditions, recent work has demonstrated that %underspecification hinders
% neural models often struggle to generalize to data outside their training domain~\citep{mccoy-etal-2019-right,Geirhos2020ShortcutLI,damour2020underspecification}.
% These models contain a huge number of parameters and thus have the freedom to identify features that perform well in the training domain but do not generalize.
% This behavior is related to the generalization difficulties reported by~\citet{harrigian-etal-2020-models}.
% Lacking information about causal relationships, neural methods are free to learn shortcuts that perform well due to dataset artifacts (e.g., control users discussing sports) but perform poorly on related datasets that do not closely follow the same construction process and thus do not contain the same shortcuts.

In this work, we explore approaches for constraining the behavior of depression detection methods by the presence of symptoms known to be related to depression, like mood and sleep issues.
To do so, we develop nine symptom detection models that correspond to questions present in PHQ9, a screening questionnaire that has been clinically validated and commonly used in practical setting~\cite{kroenke2001phq}. These questions ask how often the patient has experienced symptoms from nine symptom groups (e.g., how often have you had ``\textit{little interest/pleasure in doing things?}'').

Grounding depression detection in a trusted diagnostic tool produces several benefits.
From the perspective of mental health professionals, the output of such model is inherently more reliable than a black-box model, because classification decisions are based on the presence of specific symptoms in specific posts that can be inspected in order to assess the quality of evidence for a diagnosis.
Further, we find this improves the model's ability to generalize, which may be due to limiting its ability to use spurious shortcuts.
%this improves the model's ability to generalize by limiting its potential to use spurious shortcuts. 
This strategy is complementary to strategies for reducing temporal and topical artifacts \citep{harrigian-etal-2020-models}.
% This strategy is complementary to \citet{harrigian-etal-2020-models}'s recommendations for reducing dataset artifacts by removing temporal and topical differences between diagnosed and control users.
% Neural methods can effectively find and utilize such artifacts %~\citep{mccoy-etal-2020-berts,damour2020underspecification} %Redundant citations
% , and removing them completely is a difficult task.

Our proposed approach consists of two simple yet effective models: a questionnaire model that detects symptoms from PHQ9 and a depression detection model.
We instantiate both with a range of methods that are progressively less constrained.
At one end of the spectrum, the questionnaire model uses only manually-defined patterns and the depression model makes classification decisions by counting how many times these patterns appear in a user's posts.
At the opposite end of the spectrum, there is no explicit questionnaire model and BERT~\citep{devlin2019bert} serves as an unconstrained depression detection model.
In between, we relax the questionnaire model by training BERT-based symptom classifiers using the manually-defined patterns, by considering symptom representations rather than counts, and by adding an extra trainable `\textit{other}' symptom.

We find that our constrained models perform competitively compared to a standard unconstrained BERT classifier when trained and evaluated on the same dataset, while additionally providing a model whose behavior can be understood in terms of relevant symptoms in specific posts.
However, \textit{dataset-transfer} evaluations demonstrate substantial degradation in BERT's effectiveness. In this setting, our constrained models outperform the unconstrained BERT and show improved generalizability, even across similar datasets.
%when the datasets considered are highly similar.

% We find that the constrained models perform well when trained and evaluated on the same dataset, though BERT performs slightly better in this scenario.
% These results show that constrained depression detection models can perform competitively with a strong BERT baseline while having an explainability advantage.
% To assess whether these approaches generalize, we conduct dataset-transfer evaluations on datasets constructed from the same social media platform.
% Despite similarities between the training and evaluation datasets, BERT's effectiveness substantially degrades in this scenario, whereas the constrained models suffer a smaller decrease in effectiveness and even outperform BERT.

% This both demonstrates the generalizability of the constrained models and suggests that BERT's superior effectiveness may in part be due to spurious shortcuts.
Our contributions are:
\textit{(1)} comprehensive pattern sets for identifying the symptoms in PHQ9 and heuristics for using them to train weakly-supervised symptom classifiers,
\textit{(2)} a range of progressively less constrained methods for performing depression detection based on these symptoms, and
\textit{(3)} an extensive evaluation of depression detection methods.
%Our code and pattern resources are available at \url{https://github.com/thongnt99/acl22-depression-phq9}.
Our implementation is available online\footnote{\url{https://github.com/thongnt99/acl22-depression-phq9}}.

\begin{figure*}[h!]
    \centering
    \includegraphics[width=0.7\linewidth]{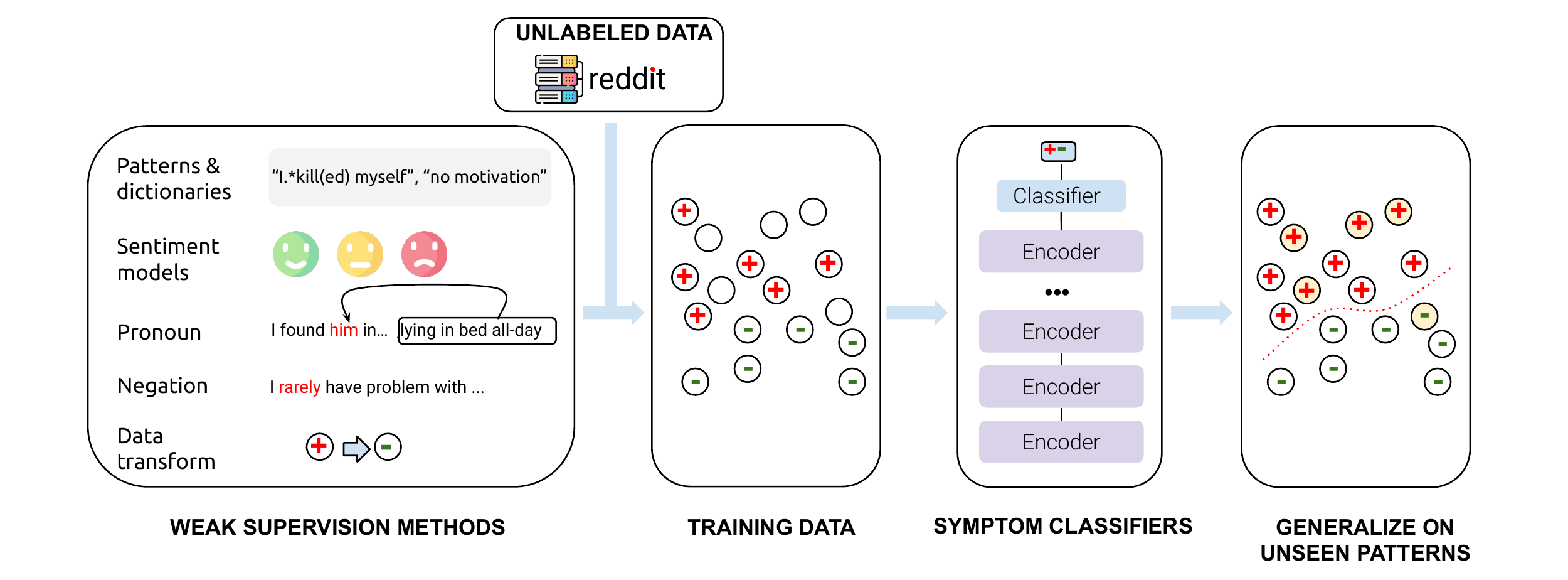}
    \caption{Weakly-supervised questionnaire model}
    \label{fig:symptom_models}
    \vspace{-2mm} 
\end{figure*}

\section{Related Work}
% In the last years, there has been a significant amount of work in applying Natural Language Processing (NLP) techniques and methods in the mental health domain. Given the sensitivity of the domain, access to clinical datasets has been a challenge for researchers. As a result, the research community has focused its efforts on social media datasets instead. For instance, the datasets that have been introduced in shared tasks have provided researchers with the ability to study mental health disorders and phenomena on large datasets compiled from Twitter and Reddit. 

Natural language processing methods have been widely used for automatic mental health assessment.
To support automated analyses of mental health related language, a variety of datasets have been proposed.
\citet{coppersmith-etal-2014-quantifying} focused on predicting depressed and PTSD users in Twitter, whereas~\citet{milne-etal-2016-clpsych,shing2018expert} and~\citet{zirikly2019clpsych} aimed to detect high risk and suicidal users from their ReachOut and Reddit posts, respectively. \citet[RSDD;][]{yates-etal-2017-depression},  \citet[SMHD;][]{cohan2018smhd}, and \citet{Wolohan2018DetectingLT} investigated identifying depression and other mental health conditions from Reddit.
% , while \citet{macavaney-etal-2018-rsdd} studied the temporal aspect of mental health language in social media. 
Rich bodies of work in this area focused on studying language use and linguistic styles in depressed users. LIWC~\cite{tausczik2010psychological} has been one of the most popular tools to characterize depression language~\cite{ramirez2008psychology,de2013predicting}.    
% Automated NLP tools have been
% able to detect mental health disorders based on language \cite{mowery-etal-2016-towards,just-etal-2019-coherence}.
Similarly to other NLP domains, the use of contextualized embeddings has improved the performance of classifiers~\cite{jiang2020detection,matero2019suicide}. 
% Use of automated NLP tools to analyze implicit mental health signals have become more critical than before, due to the ever increasing socioeconomic barriers of in-person clinical visits \cite{harrigian-etal-2020-models,bojdani2020covid}.

Recent work shows that while such NLP models achieve promising results, they have poor generalization to new data platforms and user groups; For example, \citet{harrigian-etal-2020-models} investigated various factors, including  sample size, class imbalance, temporal misalignment (e.g., language dynamic, linguistic norms), deployment latency, and self-disclosure bias that may cause performance degradation when a model is transferred to a new dataset or domain.
The issues can occur even when datasets appear to be similar, such as when Reddit-based datasets employ different rules for selecting diagnosed and control users.
% This is why computational methods for domain adaptation have been relatively well studied in NLP \citep[e.g., see][]{chu-wang-2018-survey,ma-etal-2019-domain}.
Another problem is the black-box nature of model predictions which is a major hurdle in deploying AI models in clinical practice \cite{mullenbach-etal-2018-explainable}.
In this work we aim at reducing this problem by proposing to ground depression assessment in a clinical questionnaire for measuring severity of depression.

Others have considered making predictions more explainable in the mental health domain. \citet{explainability-deep-amini-2020} focused on leveraging a user-level attention mechanism for detecting signs of anorexia in social media profiles. Our method differs from theirs in that the explanations are the results of the analysis of the attention weights, while our approaches ground model predictions in a well-established clinical instrument.
%Further, there are many caveats in using attention weights for explainability. \cite{jain2019attention}. 

Outside of our work, we are aware of two datasets that incorporate questionnaire information such as PHQ9 for identifying depression.
% eRisk 2017 Pilot task
% eRisk 2018 T1: Early Detection of Signs of Depression. This is a continuation of the eRisk 2017 pilot task; based on social media postings of depressed vs control users; no BDI
% eRisk 2019 T3: Measuring the severity of the signs of depression. This is a new task in 2019. The task consists of estimating the level of depression from a thread of user submissions. For each user, the participants will be given a history of postings and the participants will have to fill a standard depression questionnaire (based on the evidence found in the history of postings).
% eRisk 2020 T2: Measuring the severity of the signs of depression. This is a continuation of eRisk 2019's T3 task.
The most recent eRisk shared task \cite{losada2019overview} relies on the Beck Depression Inventory (BDI), a 21-item questionnaire that assesses level of depression based on the presence of feelings like sadness, pessimism, etc. Models are built to estimate the user-level BDI score at given time frames. Our approach differs in that we use PHQ9 and evaluate item scores at the post level, which grounds predictions in the presence of clinically-relevant symptoms. In eRisk, a sum of BDI scores is the modeled outcome (corresponding to our baseline pattern-based (threshold) classifier). We use user-level labels for evaluating depression status and evaluate how constraining on PHQ9 symptoms affects the user-level classification performance.
%In addition, we are also the first to study the generalizability of this approach. 

\citet{delahunty-etal-2019-passive} used a deep neural network to predict PHQ4 scores %a shorter version of PHQ9
using clinical data that contains patients' PHQ4 scores~\cite{gratch2014distress}. Our work does not require access to PHQ labeled clinical data, which can be hard to obtain at scale. Furthermore, Delahunty's approach generalizes poorly to social media data.
\citet{rinaldi2020predicting} predict depression based on screening interviews that rely on PHQ9 categories. In their setting, PHQ9 is a channel to retrieve the depressed label, but is not used for explainability. \citet{yadav-etal-2020-identifying} propose a multitask learning framework that uses PHQ-9 and figurative language detection as auxiliary tasks.
\citet{lee-etal-2021-micromodels-efficient} contemporaneously propose a \textit{micromodel} architecture that they apply to mental health assessment tasks.
Our work shares several similarities with this approach, which uses micromodels that are similar to our symptom classifiers (questionnaire models).
% \citeauthor{lee-etal-2021-micromodels-efficient} that proposes micromodel architecture for various mental health assessment tasks. The micromodels used in this work is similar to our symptom classifiers. 

\section{Methodology}
\subsection{Pattern-based methods} \label{pbc}
Our most straightforward and constrained methods are pattern-based classifiers that make classification decisions based on the presence of positive symptom patterns. This method could be decomposed into two components: a questionnaire model and a depression model. \\
\paragraphHdNospace{Questionnaire model.} The questionnaire model of pattern-based methods is simply a pattern matcher that matches each user post against symptom patterns. It produces a binary pattern matching matrix of size $(num\_post \times 9)$ whose entry at $(i, j)$ is $1$ if a match is found between the $i^{th}$ post and any pattern of the $j^{th}$ symptom (question).  \\
\paragraphHdNospace{Depression model.} We implement two variations of the depression model whose input is the pattern matching matrix generated by the previous questionnaire model: a count-based approach and a CNN approach. The count-based approach simply considers whether the number of patterns found in the pattern matrix exceeds a threshold. The CNN approach applies CNN kernels cascaded with a linear layer over the pattern matrix. This approach allows consecutive posts to be weighted differently.
In pilot experiments, we also tested a variant that closely mirrors PHQ9 by summing scores over a two-week window; this variant performed worse due to the new temporal requirement that often creates data sparsity within windows.
%, and we omit further discussion for brevity.

\subsection{Classifier-constrained methods} \label{qbm}
\subsubsection{\textbf{PHQ9}} \label{phq9} 
One drawback of pattern-based classifiers is the inflexibility of pattern matching.
The classifier-constrained methods relax the pattern-matching requirement by training a questionnaire model on the weakly-supervised data described in Section~\ref{subsection:questionnaire_dataset}.
This results in models that remain grounded in the clinical questionnaire but are capable of generalizing beyond the pattern sets. The PHQ9 architecture is  also comprised of a questionnaire model and a depression model.
% The encoder is responsible for converting raw textual posts into embedding vectors. We utilize pre-trained BERT~\cite{devlin2018bert} for this component.

\paragraphHd{Questionnaire model.}
The questionnaire model receives BERT~\cite{devlin-etal-2019-bert} token embeddings of every post and is trained to predict the answer (positive or negative) for each of the questions in the PHQ9 instrument. This model consists of $9$ symptom classifiers, \textit{anhedonia}, \textit{concentration}, \textit{eating}, \textit{fatigue}, \textit{mood}, \textit{psychomotor}, \textit{self-esteem}, \textit{self-harm} and \textit{sleep}, corresponding to the questionnaire's $9$ questions. Each symptom classifier is a CNN classifier with a linear layer on top. As illustrated in Figure \ref{fig:symptom_models}, all symptom classifiers were separately trained on weakly-labeled data, which we describe in Section \ref{subsection:questionnaire_dataset}. The questionnaire model's ability to generalize to unseen patterns comes from two sources: BERT embeddings and weakly-labeled symptom data.
First, BERT embeddings have been successfully used to transfer knowledge across domains in many NLP applications \cite{rietzler2020adapt,peng2019transfer,houlsbyparameter}. Second, in weakly-labeled data, the background or contextual text around the matched patterns could provide relevant cues, which is a means of generalization. For example, in the text ``\textit{now \underline{I don't want to do anything}. I can't do more than sleep, eat, and watch tv.}'', background phrases, such as \textit{``I can't do more...''}, are as useful as the underlined pattern for identifying the symptom \textit{anhedonia}. 

\paragraphHd{Depression model.}
The depression model predicts whether a user is depressed based on the questionnaire model's output for each post. The questionnaire model's output can be either the final question scores (i.e., symptom scores) or the hidden layers (i.e., symptom vectors) of the $9$ sub-models.
The former represents each post with a single vector of size $9$, which is compact but less informative, while the latter is a larger matrix of size $hidden\_size \times 9$ that preserves more information.
Any classification architecture could be used for this depression model. For simplicity, we use a linear classifier on top of features extracted by CNN kernels of various sizes. CNN kernels help summarize symptoms within a sliding window of consecutive posts sorted by timestamp and therefore are a relaxation of the two-week windows considered by the PHQ9 instrument. This relaxation allows more posts to be considered by each CNN kernel, which mitigates the data sparsity problem of the hard two-week window approach.
    
This depression model is trained using user-level depression labels, and while this model is being trained, the encoder and questionnaire components are frozen.
The frozen weights ensure that each questionnaire model does not drift away from its original purpose of detecting symptoms.

\subsubsection{\textbf{PHQ9Plus}} PHQ9Plus extends the PHQ9 method by appending an additional symptom (neuron) to the PHQ9 symptoms that form the questionnaire model. This neuron is connected to post embedding and produces a score for every post. Furthermore, we make this additional neuron trainable end-to-end to learn other signals similarly to PHQ9 symptoms. Doing that allows PHQ9Plus to learn from training data other depressive signals in addition to the PHQ9 symptoms. However, in return, it also risks incorporating undesirable shortcuts that harm the model's generalizability.
\subsection{Unconstrained method (BERT baseline)} \label{unconstrained_md}
In the previous classifier-constrained methods, depression classifications are constrained by a questionnaire model that determines the presence of symptoms. This is an information bottleneck intended to make the model generalize better.
% , but it may also harm performance by limiting the signals that can be considered.
In order to quantify the impact of this bottleneck, we also consider an unconstrained model that replace the questionnaire model in the previous methods by only a BERT encoder.
This gives a loose upper bound on the classifier-constrained methods's performance since this approach has access to the raw BERT embeddings and thus can utilize more signals (even spurious ones) than those captured by the questionnaire model.
% For the depression component, we used CNN with k-max pooling.

\section{Experiments}
We conduct experiments on three datasets; all consist of Reddit social media data but follow different construction methodologies (e.g., identifying depressed users based on a self-report statement vs. based on starting a thread in a support subreddit).
In addition to evaluating methods on each dataset, \textit{dataset-transfer} evaluation allows us to evaluate how well methods generalize to similar datasets with different construction methodologies.
\subsection{Depression datasets}
The three datasets selected for experiments are RSDD~\cite{yates-etal-2017-depression}, eRisk2018~\cite{losada2016test} and TRT~\cite{Wolohan2018DetectingLT}. The RSDD (Reddit Self-reported Depression Diagnosis) dataset was constructed from Reddit posts and contains approximately $9,000$ self-reported diagnosed users and $107,000$ matched control users.
%For details about the construction methodology, please refer to the cited paper.

eRisk2018 is a smaller dataset of $214$ depressed users and $1,493$ control users curated to evaluate the effectiveness of early risk detection on the Internet. Similar to RSDD, the depression group in eRisk2018 was collected based on user self-reports; Here, posts from mental health subreddits were not excluded like in RSDD. Due to the small size of the original training set, which makes the deep learning approaches unstable, we re-partitioned this dataset to allow more data for training.

Different from RSDD and eRisk2018, TRT (Topic-Restricted-Text) was constructed based on community participation. Specifically, the depressed users were drawn from members of the \textit{/r/depression} subreddit, and control users were sampled from the \textit{/r/AskReddit} subreddit. Following the construction guideline described in~\cite{Wolohan2018DetectingLT} and discussion with the authors, we re-generated a version of TRT containing $6,805$ depressed users and $57,155$ control users. 

On all datasets, we report the F1 score of the positive (i.e., diagnosed) class, and the area under the receiver operating characteristic curve (AUC). 
% AUC is our preferred metric because it measures the model's performance regardless of the decision threshold, whereas the other metrics require a decision threshold which is set to $0.5$ in this work.
\subsection{Questionnaire dataset}
\label{subsection:questionnaire_dataset}
The questionnaire model is tasked with classifying if a given post contains a PHQ9 symptom (positive) or not (negative). 
% The depression model then takes as input either the symptom score (probability) or the symptom vector used to compute the depression score.
Given the lack of training data for this task,  we collected regular expression patterns and heuristics to construct weakly-supervised training data for each of the symptoms.
We describe the process succinctly here and provide additional details in the Appendix.
We note that this weakly-supervised data is used only for training.
% This section describes the data creation process for the questionnaire models, which consists of 9 symptom detectors corresponding to 9 questions in the PHQ9 questionnaire. 

% , we collected regular expression patterns and used these patterns together with heuristics to construct training data (positive class and negative class) for each symptom. 
% \subsection{Pattern-based generated data}
\subsubsection{Positive class} \label{positive_class_data}
For each question, we prepare a set of positive symptom patterns (e.g., ``\textit{can'?t sleep}''). 
% The number of patterns for each symptom is shown in Table \ref{tab:pattern_stat}. 
Each pattern set is then matched against a post collection crawled from 127 mental-health subreddits\footnote{\url{https://files.pushshift.io/reddit/}}. In addition, we also include posts from the SMHD dataset \cite{cohan2018smhd}, which excludes posts from mental-health subreddits, to diversify the training data. 
% The purpose of using these two raw datasets is to increase the diversity and minimize the bias of the data.
In the labeling step, we select posts containing symptom patterns as positive examples.

% \begin{table}[t]
%     \small
%     \begin{tabular}{lrlr}
%     \toprule
%     \textbf{Symptom} & \textbf{\# patterns} & \textbf{Symptom} & \textbf{\# patterns} \\
%     \cmidrule(lr){1-2} \cmidrule(lr){3-4}
%     Anhedonia & 116 & Psychomotor & 108 \\
%     Concentration & 70 & Self-esteem & 102 \\
%     Eating & 145 & Self-harm & 126\\
%     Fatigue & 73 & Sleep & 118\\ 
%     Mood & 110\\ 
%     \bottomrule 
%     \end{tabular}   
%     \caption{Number of patterns for each question}
%     \label{tab:pattern_stat}
%     \vspace{-5mm}
% \end{table}

While being fast and transparent, pattern matching may produce many false positives (FPs).
We used additional heuristics to remove instances of the four most common types of FPs we observed:
%and the techniques we employed to mitigate them when constructing weakly-labeled training data.

\paragraphHdNospace{Positive sentiment.} Posts containing symptom patterns but conveying a positive/happy sentiment.

% for example, ``\textit{Friends and I stayed up all night playing a game}'' contains the positive pattern ``\textit{stayed up all night}''.
% , but it shows excitement about the game rather than a sleep issue. As a solution, we removed all posts containing positive sentiment with the help of Allen NLP's sentiment model~\cite{Gardner2017ADS}.  

\paragraphHdNospace{Conditional clause.} Posts describing a symptom hypothesis rather than an experience. 
% Sometimes users hypothesize about their health conditions, such as ``\textit{If I lost my appetite for days at a time, that...  wouldn't be sustainable for me.}''.
% We remove these posts by using regular expressions to identify popular conditional clause formats.

\paragraphHdNospace{Third-person pronouns.} Posts discussing symptoms of other people (e.g., friends, relatives) rather symptoms the user is experiencing.
% Users may attribute a condition to someone else, such as in ``\textit{he is easily distracted.}''
% We identified posts of this kind by checking whether the closest pronoun to the positive pattern is third-person or first-person, and removing posts in the former category.
% This includes references like ``\textit{my son}''.
%similar to the above example by checking whether the pronoun (``\textit{My son}'') closest to the positive pattern (``\textit{easily distracted}'') is third-person or first-person pronoun. 

\paragraphHdNospace{Negation.} Posts containing symptom patterns with negation words (e.g., ``\textit{not}'', ``\textit{never}'') preceding.
% Positive patterns may be negated, such as in ``\textit{haven't had a suicidal thought in ages.}''. To handle this situation, we removed posts containing positive patterns preceded by a negation word, such as (``\textit{not}'', ``\textit{never}'', `\textit{rarely}'', etc.). 

% While these heuristics removed many false positives, there are still harder cases that require semantic understanding of the text.
% For example, ``\textit{it's \underline{difficult to focus} on my breathing for very long}'' and  ``\textit{that behavior makes me \underline{lose my appetite}}''. We leave these difficult cases for future work.

\subsubsection{Negative class}
Identifying hard negative samples is crucial for the quality of the trained classifiers.
% Models trained on negative examples that are too easy might be prone to over-fitting or perform only keyword matching. 
We use five heuristics to identify and synthesize negative examples for each symptom:

\paragraphHdNospace{Keywords.} Posts that contain keywords (e.g., ``\textit{sleep}``) related to positive patterns (e.g., ``\textit{can't sleep}``) but do not match any positive pattern. 
% This hinders classifiers to solely use keywords to overfit the training data.
% This hinders models to perform simple keyword matching.
%The resulted posts are labeled as negative. Doing this provides harder negative examples, which in turn would hinder the trained model from doing keyword matching. 
 
\paragraphHdNospace{Pronouns.} Posts synthesized by replacing first-person pronouns (e.g.,``\textit{I}``) from the positive examples with third-person pronouns (e.g., ``\textit{She}``). 
% We replace the first-person pronouns appearing in posts from the positive class with third-person pronouns or proper nouns, such as replacing ``\textit{I}'' with ``\textit{she}.''
%This method could help further removing the FP case about third-person pronoun described in Section \ref{positive_class_data}. 

\paragraphHdNospace{Other symptoms.} Posts sampled randomly from positive examples of other symptoms (without matching a pattern for the current symptom).
% We use randomly selected posts labeled positive for other questions (symptoms) as negative examples for the given question.
% We ensure the selected posts do not contain positive patterns for the given question.
%positive class from other questions could also be used as negative examples of the considered question. We also have to make sure that the selected posts do not contain positive patterns of the current question. 

\paragraphHdNospace{Negation.} Posts synthesized by negating symptom patterns in positive examples using hand-crafted mappings (e.g., ``\textit{tired}`` to ``\textit{never tired}``).

% For each positive pattern defined in the previous section, we created a corresponding negated one, such as negating ``\textit{have sleep apnea}'' to ``\textit{never have sleep apnea}''.
%Given this mapping, we loop through all sentences in posts of positive class, select those containing a positive pattern in the mapping, negate and use them as negative samples.
% Only matched sentences were used in this method, because using the whole post with only some sentences being negated could lead to contextual inconsistencies.
%, which in turn harming the trained classifier.  

\paragraphHdNospace{Positive sentiment.} Posts sampled from neutral or positive classes in the Sentiment140 sentiment analysis corpus \cite{go2009twitter}. 
% In particular, we used the Sentiment140 corpus, which contains 1.6 millions tweets .
\subsection{Experimental setup}
We designed experiments to analyze our two main component: the questionnaire and depression models. The setup for these experiments is summarized in Table \ref{tab:exp_variations} and specific hyperparameters are described in the Appendix. 

% The hyper-parameters of those models were set as follows:

% \textbf{Questionnaire model}: In questionnaire-based methods, we trained a CNN for each of the nine symptom classifiers. We used filters of sizes $[2, 3, 4, 5, 6]$, and one filter for each size. Consequently, the max-pooling produces a vector of size $5$, which is then fed into the final linear layer for prediction. We apply $L_2$ regularization specifically to the CNN kernels. The $L_2$ weights were fine-tuned with three options [$0.1$, $0.01$, $0.001$]. 

% \textbf{Depression model}: We experimented with $6$ variations described in Table \ref{tab:exp_variations}. 
% % These correspond to Sections \ref{pbc}, \ref{qbm}, and \ref{unconstrained_md}, respectively.
% All of these variations, except for the first one, use a CNN classifier on top of different inputs ranging from BERT embedding to pattern scores. The CNN used here has ($filter\_size = [2,3,4,5,6]$, $num\_filter = 50$), and $(k=5)$ for k-max pooling. The threshold for the first variation was tuned on from $1$ to $10$. 

% In all experiments with pre-trained BERT, we used the BERT-base version \cite{devlin2018bert}. We do not fine-tune BERT's parameters since in a pilot study, we found that fine-tuning BERT does not improve the generalization. All models were optimized using a cross-entropy loss with class weights of $0.1$ and $0.9$ for the control and depressed classes, respectively, a 1cycle learning rate scheduler (with a maximum of $0.01$), batches of size 64, and early stopping after five epochs.
\begin{table}[h!]
    \centering
    \small 
    \resizebox{\linewidth}{!}{
    \begin{tabular}{@{}llll@{}}
    \toprule
    \textbf{Method} & \textbf{Encoder} & \textbf{Symptom REP(*)} & \textbf{DM(*)}\\
    \midrule
    Pattern (threshold) & -  & Pattern matrix & - \\
    Pattern (CNN) & - & Pattern matrix & CNN \\
    PHQ9 (scores) & BERT  & Scores & CNN\\
    PHQ9 (vectors) & BERT  & Vectors & CNN\\
    PHQ9Plus & BERT  & Scores + other & CNN\\
    Unconstrained (BERT) & BERT  & - & CNN  \\ 
    \bottomrule
    \end{tabular}
    } % end resizebox
    \caption{Experimental variations. (*) REP: representation; DM: Depression model}
    \label{tab:exp_variations}
    \vspace{-5mm} 
\end{table}
\begin{figure*}
    \centering
    \includegraphics[scale=0.4]{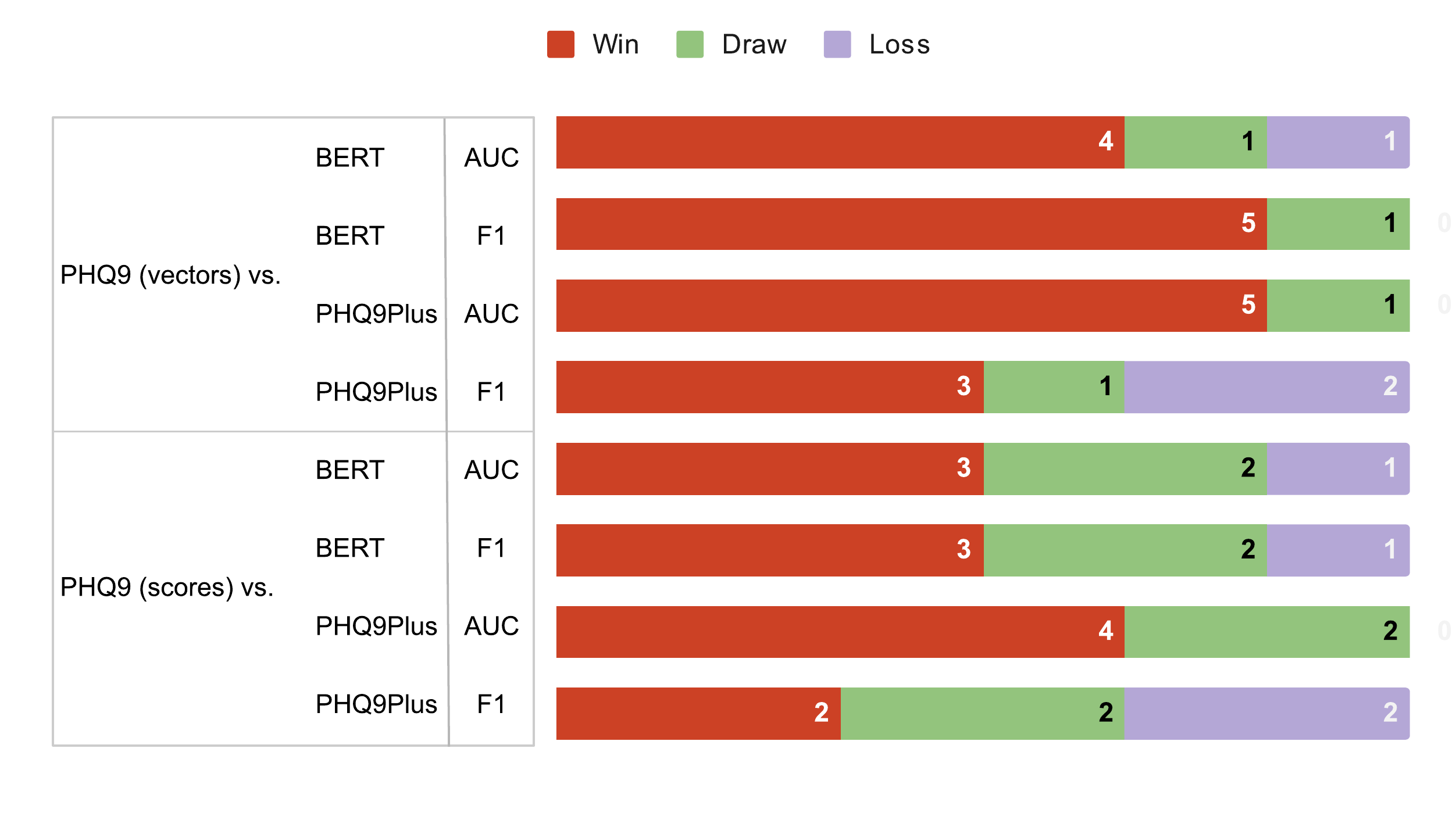}
        \vspace{-2mm}
        \caption{Relative comparison between PHQ9 methods vs. BERT and PHQ9Plus in \textit{dataset-transfer} settings. Win (or Loss): PHQ9 performs significantly better (or worse);  Draw: not significant different. (T-test, $\alpha = 0.05$)}
    \label{fig:relative_comparision}

\end{figure*}
\begin{table*}[ht!]
    \centering
    \small
    \resizebox{\linewidth}{!}{%
    \begin{tabular}{@{}llcccccc@{}}
    \toprule
    \multirow{3}{*}{\textbf{Train}} &
    \multirow{3}{*}{\textbf{Method}} &  \multicolumn{2}{c}{\textbf{Test: eRisk}} & \multicolumn{2}{c}{\textbf{Test: RSDD}} & \multicolumn{2}{c}{\textbf{Test: TRT}} \\
    \cmidrule(lr){3-4} \cmidrule(lr){5-6} \cmidrule(lr){7-8}
    &  &  AUC  &  F1  &   AUC  &  F1  &  AUC  &  F1  \\
    \midrule
    \multirow{7}{*}{\textbf{TRT}} 
    & LIWC+ngram~\cite{Wolohan2018DetectingLT} &  -  &  -  &  -  &  -  &  \color{gray} $0.79\pm***$  &  \color{gray} $0.73\pm***$ \\
    & Pattern (threshold) &  -  &  0.38$\pm$0.00  &  -  &  \textbf{0.35$\pm$0.00}  &  \color{gray} -   &  \color{gray} 0.46$\pm$0.00  \\ % threshold=2
    & Pattern (CNN) &  0.79$\pm$0.01  &  0.40$\pm$0.02  &  0.71$\pm$0.00  &  0.26$\pm$0.01  &  \color{gray} 0.80$\pm$0.01  &  \color{gray} 0.51$\pm$0.02 \\
    & PHQ9 (scores) &  0.85$\pm$0.01  &  \textbf{0.41$\pm$0.01}  & \textbf{0.78$\pm$0.03}  &  \textbf{0.35$\pm$0.03}  &  \color{gray} 0.92$\pm$0.01  &  \color{gray} 0.64$\pm$0.02  \\ 
    & PHQ9 (vectors) &  \textbf{0.86$\pm$0.00}  &  0.31$\pm$0.01  &  0.73$\pm$0.00  &  0.31$\pm$0.00  &  \color{gray} 0.96$\pm$0.00  &  \color{gray} 0.77$\pm$0.00 \\
    & PHQ9Plus &  0.80$\pm$0.01  &  0.40$\pm$0.07  &  0.59$\pm$0.03  &  0.21$\pm$0.02  &  \color{gray} 0.95$\pm$0.00  &  \color{gray} 0.79$\pm$0.00  \\ 
    & Unconstrained (BERT) &  0.84$\pm$0.01  &  0.15$\pm$0.02  &  0$.66\pm0.03$  &  0.22$\pm$0.03  &  \color{gray} \textbf{0.98$\pm$0.00}  &  \color{gray} \textbf{0.82$\pm$0.00} \\
    \midrule
    \multirow{7}{*}{\textbf{RSDD}}
    & CNN(400)~\cite{yates-etal-2017-depression} &  -  &  -  &  \color{gray} -  &   \color{gray} $0.51\pm***$  &  -  &  - \\ 
    & Pattern (threshold) &  -  &  0.38$\pm$0.00  &  \color{gray} -  &  \color{gray} 0.35$\pm$0.00  &  -  &  0.46$\pm$0.00  \\
    & Pattern (CNN) &  0.79$\pm$0.01  &  0.47$\pm$0.00  &  \color{gray} 0.74$\pm$0.01   &  \color{gray} 0.36$\pm$0.02  &  0.79$\pm$0.00  &  0.39$\pm$0.01  \\
    & PHQ9 (scores) &  0.80$\pm$0.01  &  0.43$\pm$0.01   &  \color{gray} 0.85$\pm$0.00  &  \color{gray} 0.47$\pm$0.01  &  0.82$\pm$0.00  &  0.46$\pm$0.00 \\
    & PHQ9 (vectors) &  0.81$\pm$0.01  &  0.46$\pm$0.01   &  \color{gray} 0.85$\pm$0.00  &  \color{gray} 0.49$\pm$0.01  &  \textbf{0.86$\pm$0.00}  &  \textbf{0.52$\pm$0.00} \\
    & PHQ9Plus &  0.81$\pm$0.03  &  \textbf{0.49$\pm$0.00}  &  \color{gray} \textbf{0.86$\pm$0.02}  &  \color{gray} \textbf{0.55$\pm$0.00}  &  0.82$\pm$0.00  &  0.49$\pm$0.00  \\
    & Unconstrained (BERT) &  \textbf{0.84$\pm$0.01}  &  0.44$\pm$0.02  &  \color{gray} \textbf{0.86$\pm$0.00}  &  \color{gray} 0.53$\pm$0.01   &  0.82$\pm$0.00  &  0.47$\pm$0.00 \\    
    \midrule
    \multirow{6}{*}{\textbf{eRisk}}
    % removed due to our rebalancing of eRisk (making these incomparable)
    %  & Glove-CNN \cite{trotzek2018word} &  \color{gray} -  &  \color{gray} $0.54\pm***$  &  -  &  -   &  -  &  - \\ 
    % & BoW Ensemble~\cite{trotzek2018word} &  \color{gray} -  &  \color{gray} $0.64\pm***$  &  -  &  -  &  -  &  - \\  
    & Pattern (threshold) &  \color{gray} -  &  \color{gray} 0.40$\pm$0.00  &  -  &  0.32$\pm$0.00  &  -  &  0.44$\pm$0.00 \\
    & Pattern (CNN) &  \color{gray} 0.80$\pm$0.00  &  \color{gray} 0.43$\pm$0.01  &  0.73$\pm$0.01  &  0.31$\pm$0.01  &  0.79$\pm$0.00   &  0.47$\pm$0.01 \\
    & PHQ9 (scores) &   \color{gray} 0.87$\pm$0.00  &  \color{gray} 0.54$\pm$0.02  &  0.81$\pm$0.01  &   0.38$\pm$0.00  &  \textbf{0.90$\pm$0.00}  &  \textbf{0.56$\pm$0.01} \\
    & PHQ9 (vectors) &  \color{gray} 0.88$\pm$0.00  &  \color{gray} 0.55$\pm$0.00  &   \textbf{0.82$\pm$0.00}  &   \textbf{0.39$\pm$0.01}  &  0.89$\pm$0.00  &  \textbf{0.56$\pm$0.04} \\
    & PHQ9Plus &  \color{gray} 0.94$\pm$0.00  &  \color{gray} \textbf{0.73$\pm$0.03}  &  0.79$\pm$0.01  &  0.35$\pm$0.01  &  0.84$\pm$0.01  &  0.54$\pm$0.02 \\ 
    & Unconstrained (BERT) &  \color{gray} \textbf{0.95$\pm$0.01}  &  \color{gray} 0.71$\pm$0.03  &  0.81$\pm$0.02  &   0.36$\pm$0.02  &   0.83$\pm$0.01  &  0.50$\pm$0.02 \\
    \bottomrule
    \end{tabular}%
    }
        \vspace{-2mm} 
    \caption{Depression detection on RSDD, eRisk and TRT datasets (first lines are prior work's results). Highest scores marked in bold.
    All of our methods and CNN(400) use only a user's first 400 posts,  while other baselines use all posts.
    Summary with statistical test in Figure \ref{fig:relative_comparision}.}
    \label{tab:main_results}
\end{table*}

\begin{table*}[t!]
    \centering
    \small
    \resizebox{\linewidth}{!}{%
    \begin{tabular}{p{0.09\textwidth}p{0.43\textwidth}p{0.43\textwidth}}
    \toprule
    \textbf{Model} & \textbf{Post 1} & \textbf{Post 2} \\ 
    \midrule
    PHQ9 (scores)  & Its too late to improve myself [...] I'll graduate soon, but I feel depressed, I'm overweight, and have low confidence and self-esteem [...] now there's nothing left but working for the rest of my life. no friends, no social life, nothing fun - just work. \newline 
    \tcbox{anhedonia} \tcbox{mood} \tcbox{self-esteem}
    &  I'm so tired of living this life [...] I just want it to end. maybe life is just so unfair and there's no explanation for why things are unfair. all the unfair things just get me frustrated [...] im overweight and I never succeed in losing weight, I fuck up every time I try [...] 
    \newline 
    \tcbox{anhedonia} \tcbox{self-harm} \tcbox{mood} \tcbox{fatigue}
    \\
\hline 
PHQ9Plus and  BERT
& 
I don't like the way my life is going [...] everyday is pretty much entirely spent in my room, except for several hours at the gym [...] That's also why I want people to like me, so that I'd have people to do cool things with and my days would be less lonely and boring. 
& 
So what's life like after college?
I have to admit that I'm scared as fuck about it. I'm afraid that there will be no time for fun or socializing, and that I'll always have to act all grown up and professional.\\ 
\bottomrule
    \end{tabular}}
        \vspace{-2mm}
    \caption{A depressed user's two most informative posts found by PHQ9 (scores), PHQ9Plus, and Unconstrained (BERT) models. All posts are paraphrased for anonymity.}
    \label{tab:case-study}
    \vspace{-2mm}
\end{table*} 

%\subsection{Results and analysis}
\subsection{Depression detection results}

The results from prior work and our methods  on RSDD, TRT, and eRisk are 
shown in Table~\ref{tab:main_results}.
%shown in Table~\ref{tab:prior_results} and Table~\ref{tab:main_results} respectively. 

Depression detection results in \textit{dataset-transfer} evaluation are shown in non-gray blocks in Table~\ref{tab:main_results}. Unlike standard within-dataset evaluations, this scenario requires methods trained on one dataset to generalize to other (\textit{highly similar}) datasets. While all datasets consider the same social media platform, their dataset construction methodologies differ, and thus, they are likely to contain different dataset artifacts. Unconstraind methods have the flexibility to learn shortcuts induced by these artifacts, which can lead to poor generalization beyond the training corpus. This effect is observed in the results of the unconstrained model: as summarized in Figure \ref{fig:relative_comparision}, BERT is outperformed by our PHQ9 (scores, vectors) or even pattern-based methods in many dataset-transfer settings.
%PHQ9 (scores) performs no worse than PHQ9Plus in terms of F1 and substantially better than PHQ9Plus as measured by AUC (four wins, two draws, no losses). 
In terms of F1 and AUC, our two PHQ9 variants generalize better than BERT with only $1$ ``Loss'' at most over $6$ \textit{dataset-transfer} settings, and the number of ``Win'' always dominates.  Compared to BERT, our PHQ9 (vectors) obtains $5$ ``Win'', $1$ ``Draw'', and no ``Loss'' in terms of F1.  Regarding AUC, we only observe 1 ``Loss'' replacing a ``Win''. The method using PHQ9 scores generalizes slightly worse than the one using vectors but still performs better than the unconstrained models. For example, when trained on TRT and tested on RSDD, our PHQ9 (scores) method improves over BERT by roughly 59\% F1 score and 18\% AUC. This behavior may reflect the unusual selection of control users in TRT, where control users are sampled from ``r/AskReddit''. This may introduce shortcuts (e.g., specific topics or styles) that make BERT vulnerable to the change of testing environment. Our classifier-constrained methods with scores and vectors are designed to avoid the spurious shortcuts present in this setting.

For similar reasons, the extra neuron gives the PHQ9Plus model more freedom to learn shortcuts, leading to inferior generalization than PHQ9 (with both scores and vectors). In addition to generalizing better, the methods with symptom scores can be used to identify evidence in the form of specific posts related to the symptoms in PHQ9, which makes them more trustworthy from the perspective of mental health professionals who can examine the posts to verify that symptoms are present.

On standard within-dataset evaluations (in \textcolor{gray}{gray} cells), when models are trained and tested on the same corpus, we find that F1 and AUC increase as the models become less constrained, with the standalone BERT model and PH9Plus performing the best on all datasets. However, as previously shown, this performance does not transfer to more realistic dataset-transfer settings. The two pattern-based methods perform worse than the best prior method on each dataset, though they are the easiest to interpret due to the PHQ9 symptom scores associated with each post.

When the patterns are used to train a PHQ9 (scores) model, both F1 and AUC increase substantially, with the largest improvement of 0.13 F1 and 0.12 AUC in TRT. Methods using  PHQ9 (vectors) perform slightly better than those using scores, but the latter is easier to interpret since each post is associated with a symptom score. Both perform well in comparison with the baselines despite the fact that they are constrained by the PHQ9 symptoms. The add-on neuron contributes significantly to the in-domain effectiveness of PHQ9Plus, which even outperforms BERT in several settings. 

\begin{table}[ht!]
        \centering
        \small
        \begin{tabular}{@{}lcccc@{}}
        \toprule
        \textbf{Question} & \textbf{\#Pos$/$Neg} & \textbf{F1} & $\bm{\kappa}_{(*)}$ \\ %%macro-avg
        \midrule 
        Anhedonia & 75/25  &  $0.54\pm0.08$ & 0.70 \\ 
        Concentration & 72/28 & $0.83\pm0.03$ & 0.91 \\
        Eating & 53/47 & $0.87\pm0.01$ & 0.68\\
        Fatigue & 46/63 & $0.62\pm0.01$ & 0.66 \\
        Mood & 57/43 &  $0.64\pm0.02$ & 0.72\\
        Psychomotor & 47/53 & $0.69\pm0.01$ & 0.80 \\
        Self-esteem & 52/48 & $0.77\pm0.02$ & 0.80\\
        Self-harm & 43/57 & $0.82\pm0.02$ & 0.80 \\
        Sleep & 68/32 & $0.64\pm0.04$ & 0.68\\
        \bottomrule 
        \end{tabular}
        \caption{F1 score on manually-labeled samples over five runs and Cohen's kappa ($\kappa$) between annotators.}
        \label{tab:manual-test}
        \vspace{-2mm} 
\end{table}

\subsection{Symptom detection results}
To quantify the performance of our weakly-supervised questionnaire (symptom) models, we additionally prepared a dataset of $900$ samples manually labeled by three annotators. The annotation procedures are described in the Appendix \ref{sec:manual_dataset}. 
% ($100$ for each symptom).
The results of our symptom classifiers evaluated on the test sets are shown in Table~\ref{tab:manual-test}. Overall, our symptom classifiers perform well despite being trained on weak labels. The ``\textit{concentration}'', ``\textit{eating}'' and ``\textit{self-harm}'' classifiers show strong performance, while a lower F1 is observed with the ``\textit{anhedonia}'', ``\textit{mood}'' and ``\textit{fatigue}'' classifiers. Interestingly, we find that the F1 scores of symptom classifiers tend to positively correlate with the annotator's agreement (Pearson $\rho > 0.5$). This suggests the low F1 score in some symptom classifiers, such as \textit{``anhedonia''} and \textit{``fatigue''}, might partly be due to the ambiguity of texts. For example, it is challenging to distinguish between an ordinary bad mood versus a depressive mood. Additionally, in our analysis, we find many wrong predictions where posts use symptom-like language in a more specific context, such as ``{\textit{I completely lost my interest in him}}'' or ``{\textit{I can't concentrate on that movie}}''. These alone might not indicate a symptom, but recurrence of them might be significant.
%In some limited cases, clues revealing symptoms appear after the first 512 tokens in a post, so they are omitted by the BERT encoders in symptom classifiers. 

%\section{Ablation studies}
\subsection{Do symptom classifiers generalize?}
\label{subsec:generalize}
% Our symptom detectors could  thanks to the pre-trained BERT embedding and background cues around the matched patterns. 
To examine whether our symptom classifiers can generalize beyond pattern matching, we split each pattern set into two non-overlapping groups (\textit{g1}, \textit{g2}), which split the original dataset into two exclusive subsets. Because pattern distribution are uneven, the resulting subsets are sometimes imbalanced.
% ; patterns in one group never occur in the dataset created by the other group. 
We then evaluate our symptom classifiers on two settings (i.e., train on g1 \& test on g2, and train on g2 \& test on g1). The results shown in Table~\ref{tab:symptom_generalization} show that our symptom classifiers still achieve fairly high F1 scores in both settings. Note that, on some symptoms (e.g., concentration, fatigue), given a small coverage of patterns in \textit{g2}, our models could still achieve good performance compared to models trained on the much larger data covered by \textit{g1}. This  suggests that the symptom classifiers can generalize beyond the specific patterns they were trained with.

\begin{table}[ht!]
    \centering
    \small
    \resizebox{\linewidth}{!}{%
    \begin{tabular}{@{}lllll@{}}
    \toprule
    Symptom & \%\textit{g1(*)} & \%\textit{g2(*)} & Train \textit{g1} & Train \textit{g2} \\ 
    & & & Test \textit{g2}  &Test \textit{g1} \\
    \midrule
     Anhedonia & 0.59 & 0.41 & $0.82\pm0.04$  & $0.79\pm0.03$ \\
     Concentraion & 0.86 & 0.14 & $0.79\pm0.03$  & $0.72\pm0.03$ \\
     Eating & 0.53 & 0.47 &$0.76\pm0.02$ & $0.72\pm0.04$\\ 
     Fatigue & 0.96 & 0.04 & $0.76\pm0.03$ & $0.71\pm0.02$ \\ 
     Mood & 0.90 & 0.10 & $0.72\pm0.02$ & $0.66\pm0.04$\\ 
     Psychomotor & 0.80 & 0.20 & $0.70\pm0.02 $ & $0.66\pm0.05$\\ 
     Self-esteem & 0.95 & 0.05 & $0.70\pm0.02$ & $0.68\pm0.02$\\ 
     Self-harm & 0.60 & 0.40 & $0.68\pm0.02$ & $0.66\pm0.01$\\ 
     Sleep & 0.59 & 0.41 & $0.86\pm0.03$ & $0.72\pm0.01$\\ 
    \bottomrule
    \end{tabular}}
    \caption{F1 evaluating symptom classifiers on different pattern sets.
    (*) The proportion of the original dataset covered by each pattern group.}
    \label{tab:symptom_generalization}
    \vspace{-4mm} 
\end{table}

\label{subsec:effect_labeling_method}
\begin{figure}[ht!]
    \centering
    \includegraphics[width=\linewidth]{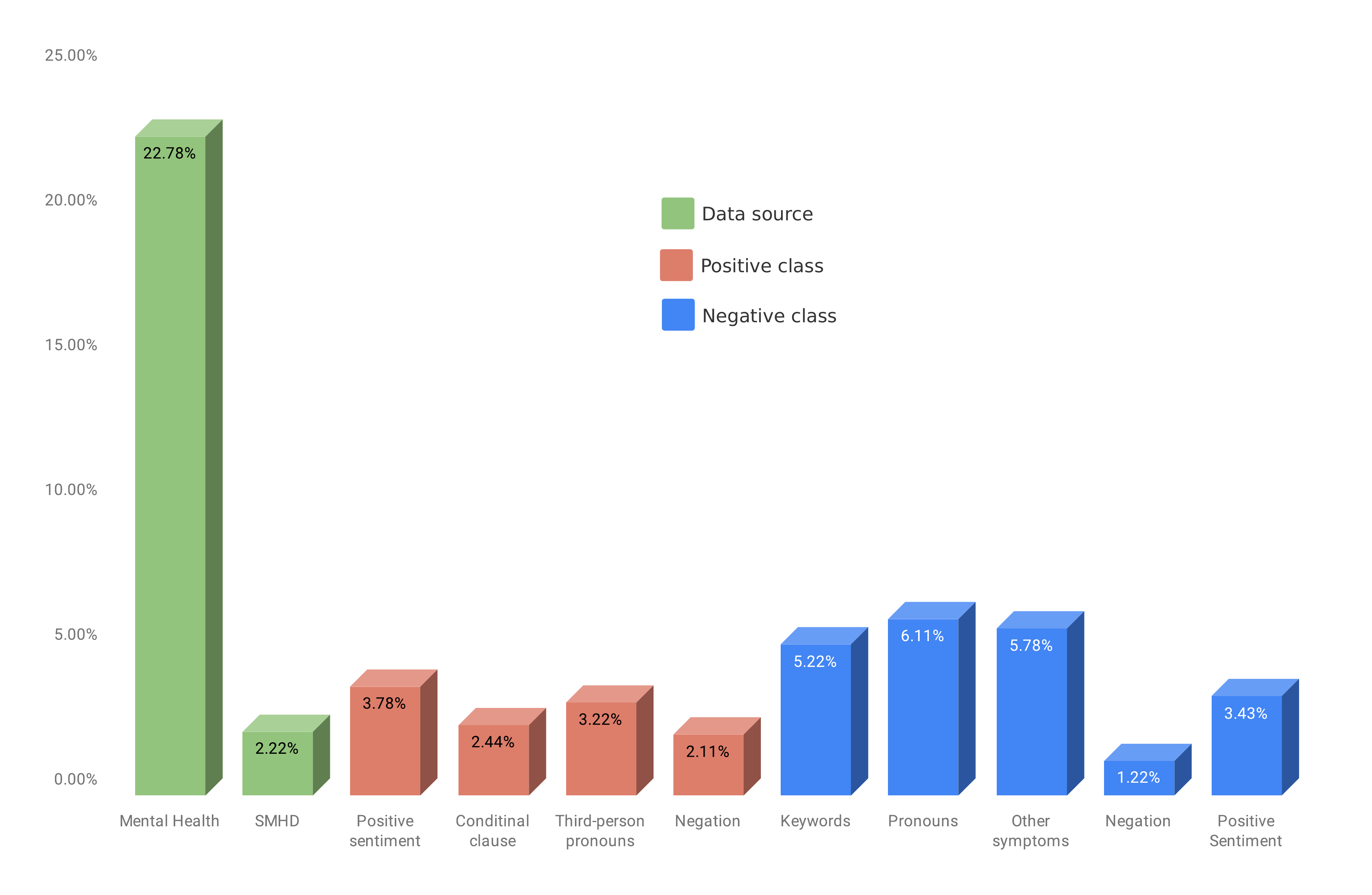}
    \caption{Effect of various factors on symptom detection. Expressed as percentage point difference in F1.}
    \label{fig:labeling_ablation}
    \vspace{-4mm}
\end{figure}

\subsection{Effect of labeling methods}
In Figure \ref{fig:labeling_ablation}, we visualize the effect of various data-construction factors on the performance of symptom classifiers. Regarding the \textit{data source}, data obtained from mental health subreddits has more influence on effectiveness than the more general posts in SMHD. Discarding data from mental health subreddits leads to an average drop of nearly 0.23 F1 score in all symptoms, while the decrease after removing SMHD is 0.02. We attribute the immense contribution of data from mental health subreddits to the fact that mental health is the main topic of discussion in those forums; therefore, pattern matching returns fewer false-positive cases and denser symptoms, resulting in better quality training data.
% In addition, the test examples are selected from the mental health community; therefore, the linguistic style may deviate from SMHD.
We further investigate the role of each method to remove FP matches in the \textit{positive class}. For that purpose, we put filtered-out FP examples back into the training data and observe the variation of F1 score on the manually labeled test sets.  In general, adding back FP examples filtered by our methods causes a total drop of nearly 0.12 in the averaged F1 score. Among them, instances with positive sentiment cause the highest decrease of roughly 0.04. Posts with third-person pronouns contribute around 0.03, while conditional clause and negation contribute modestly at around 0.02 F1.
Similarly, we analyze the effectiveness of methods to weakly annotate the \textit{negative class} by removing each of them from the training data and record the change in F1 score. We find that removing three methods, including keywords, pronouns, and other symptoms, causes a similar drop of roughly 0.06 each. Interestingly, eliminating data with positive sentiment from the \textit{negative class} has a similar effect to adding them to the \textit{positive class}, causing a drop of almost 0.04 F1. The method that changes positive examples to negative examples has the smallest impact on the F1 score (roughly 0.01).
Overall, except for the data sources, no single labeling method has a superior impact on the quality of symptom classifiers than other methods.
%Thus, our approach of combining multiple techniques is needed for good performance. 

\subsection{Contribution of PHQ9 symptoms}
\label{subsec:symptom_contribution}
To measure the contribution of a symptom to detecting depression, we remove the corresponding symptom from the model and observe the drop in the F1 score. The results are reported in Table \ref{tab:symptom_contribution}. On average, we could see that \textit{``self-harm''}, \textit{``fatigue''}, and \textit{``anhedonia''} are the strongest indicators of depression. Removing them causes a 0.13-0.17 drop in the F1 score. This is in line with the prior finding that suicidal ideation or self-harm is highly correlated with depression~\cite{braadvik2018suicide}. \textit{``Mood''}, \textit{``psychomotor''}, and \textit{``self-esteem''} contribute moderately to depression detection, with roughly a 0.09 drop in F1 score for each. The remaining three symptoms, including concentration, eating, and sleep, play a less important role in detecting depression, with each contributing around 0.05 to the F1 score.
\begin{table}[ht!]
    \centering
    \small
    \resizebox{\linewidth}{!}{%
    \begin{tabular}{@{}lclc@{}}
    \toprule
    \textbf{Symptom} & \textbf{Contribution} & \textbf{Symptom} & \textbf{Contribution} \\
    \cmidrule(lr){1-2} \cmidrule(lr){3-4}
    Anhedonia & 0.13 & Psychomotor & 0.10 \\
    Concentration & 0.05 & Self-esteem & 0.09  \\
    Eating &  0.05 & Self-harm & 0.17\\
    Fatigue & 0.13 & Sleep & 0.04 \\
    Mood & 0.09 & &  \\
    \bottomrule
    \end{tabular}}
    \caption{Contribution of symptoms to depression detection.} 
    \label{tab:symptom_contribution}
    \vspace{-4mm}
\end{table}

\subsection{Comparison with Few-shot Learner}
\label{subsec:compare_gpt-3}
Recent work has demonstrated that GPT-3 is a strong few-shot learner \cite{NEURIPS2020_1457c0d6}. Herein, we are interested in how well our classifier-constrained methods compare to the GPT-3 with prompted examples. 
We prompt GPT-3 with four examples for each (\textit{positive}, \textit{negative}) class from one dataset (e.g., TRT) and evaluate on other datasets (e.g., RSDD, eRisk). Due to the high computational cost of GPT-3, we only evaluate on $100$ positive samples and $100$ negative samples from each dataset.

We can see in Table \ref{tab:gpt3_compare} that prompted GPT-3 is consistently outperformed by our classifier-constrained methods, and the margin is often large. For example, among models trained on RSDD, the classifier-constrained model with CNN vectors achieves the highest F1 of 0.79 and 0.64 when tested on TRT and eRisk, respectively. GPT-3 performs worse with at least a 0.12 drop in F1. This result demonstrates that depression detection is still challenging for large few-shot learners, further highlighting our contributions of generalizable methods.
However, we note that this setting has several limitations that prevent a completely fair comparison.
Our methods have access to hundreds of posts, while GPT-3 has a limitation on the prompt length. In addition, prompt examples, which have high influence on the GPT-3 few-shot performance, need to be carefully selected and tuned. It is possible that we were unable to identify near-optimal prompts.
Furthermore, it is difficult to know which posts or users should be prompted to GPT-3, so we opted to select randomly. Lifting this limitation would require a separate model to identify which posts should be used as input.

\begin{table}[ht!]
\small
\centering
\begin{tabular}{@{}lcc@{}}
\toprule
\textbf{Prompt/Train RSDD} & \textbf{Test TRT} & \textbf{Test eRisk} \\
\midrule
PHQ9 (scores) & 0.64 & 0.62 \\
PHQ9 (vectors) & \textbf{0.79} & \textbf{0.64} \\
GPT-3 & 0.59 & 0.52 \\
\toprule
\textbf{Prompt/Train TRT} & \textbf{Test RSDD} & \textbf{Test eRisk} \\
\midrule
PHQ9 (scores) & \textbf{0.71} & 0.72 \\
PHQ9 (vectors) & 0.69 & \textbf{0.78} \\
GPT-3 & 0.61 & 0.54 \\
\toprule
\textbf{Prompt/Train eRisk} & \textbf{Test RSDD} & \textbf{Test TRT} \\
\midrule
PHQ9 (scores) & \textbf{0.85} & \textbf{0.74} \\
PHQ9 (vectors) & 0.83 & 0.71 \\
GPT-3 & 0.54 & 0.49\\
\bottomrule
\end{tabular}
\caption{F1 scores of PHQ9 models vs. GPT-3}
\label{tab:gpt3_compare}
\vspace{-4mm} 
\end{table} 

\section{Case study}
%  Though all models predict correctly, our PHQ9 (scores) model found more relevant posts with convincing associated symptoms
In Table \ref{tab:case-study}, we demonstrate approaches trained on TRT using text from an anonymized and paraphrased depressed user from the eRisk2018 dataset.
We show the top two posts ranked by the drop in depression score when excluding each post. All  models were able to produce correct labels with very high confidence. However, there is a clear difference in the posts that models rely primarily on for prediction. The PHQ9 (scores) model found highly relevant posts with convincing associated symptoms. For example, in the first post, the PHQ9 models found 3 symptoms, including ``\textit{anhedonia}'', ``\textit{mood}'' and ``\textit{self-esteem}''. By looking at those posts and symptoms, mental health professionals could quickly understand the patient's circumstances and make further decisions.
%This demonstrates that grounding the approach in known-relevant symptoms 
%This demonstrates the auditability of our PHQ9 models.
The two most important posts for PHQ9Plus and BERT are more about daily life concerns or complaints, which may be less useful to explain a high  depression score than the top posts used to explain the PHQ9 (scores) model.
While these posts are relevant, they are more difficult to interpret than posts directly mentioning symptoms that are known to be relevant. 
Furthermore, in the TRT training dataset, due to the biased selection of control users, those life concerns/complaints may form a shortcut that effectively differentiates depressed users from control users. However, in more realistic deployment scenarios (i.e. \textit{dataset-transfer} settings), the fact that such shortcuts do not generalize makes PHQ9Plus and BERT more unreliable and fragile. 

\section{Conclusion}
In this work, we propose a spectrum of methods for depression detection that are constrained by the presence of PHQ9 symptoms.
In our experiments on the three datasets, we find these methods to perform well compared to strong baselines while generalizing better to similar datasets. 
This can be viewed as a proof-of-concept demonstrating that grounding depression predictions in PHQ9 can improve the generalizability of depression detection and the interpretability of the model.
While this research focuses only on depression detection, the idea of constraining models to consider only relevant causes may be applied to a wider range of tasks, including detection of other mental health conditions with diagnostic questionnaires.
%In \textbf{Appendix} \ref{sec:appendix}, we additionally analyze the contribution of PHQ9 symptoms to depression detection and also show that our classifier-constrained methods outperformed GPT-3 significantly.

% On the down side, we acknowledge that our pattern sets and our symptom manual labels were created by authors who are non-mental health processionals. We attempted to increase the credibility by cross-checking among authors, and we also reported a high agreement score between annotators.
% This work should also be viewed as a proof-of-concept shows that grounding on PHQ9 symptoms would improve the generalizabilty and explainability of depression detection. To turn this proof-of-concept into a production, we should re-iterate the whole process with the involvement of clinicians. We decided to leave this for our future work. 

% \section*{Ethics}
\newpage  
\section*{Ethics Statement}
% and \citet{suster-etal-2017-short}
Due to the sensitivity of the mental health related data, additional consideration needs to be taken into account when accessing and analyzing such data, as highlighted by \citet{benton-etal-2017-ethical}.
All datasets used in this research were obtained according to each dataset's respective data usage policy. We did not interact with users in any way, and we refrained from showing any direct excerpts of the data in this manuscript to prevent risks from identifying users' pseudonyms. (All excerpts have been paraphrased.) Similarly, we made no attempt to identify, deanonymize, or link users to other social media accounts.
These precautions ensure we do not draw attention to specific users who may be suffering from depression.

All models proposed in this research were trained on social media data. Thus, they are likely to fail on data coming from other sources (e.g., clinical notes), and there are no accuracy guarantees even within social media data. Our models are not intended to replace clinicians. Instead, we envision the approaches we describe being used as assistive tools by mental health professionals.
%This machine-doctor collaboration also helps controlling any risks caused by model failures on mental health patients.     
% We note that all data used in our work is anonymized and publicly available.
% In the US, we agree it would be found exempt under 45 CFR 46.104(d)(2) given that all data is anonymized and publicly available. https://www.hhs.gov/ohrp/regulations-and-policy/decision-charts-2018/
\bibliography{anthology,custom}

\begin{thebibliography}{40}
\expandafter\ifx\csname natexlab\endcsname\relax\def\natexlab#1{#1}\fi

\bibitem[{Amini and Kosseim(2020)}]{explainability-deep-amini-2020}
Hessam Amini and Leila Kosseim. 2020.
\newblock Towards explainability in using deep learning for the detection of
  anorexia in social media.
\newblock In \emph{Natural Language Processing and Information Systems}, pages
  225--235, Cham. Springer International Publishing.

\bibitem[{Benton et~al.(2017)Benton, Coppersmith, and
  Dredze}]{benton-etal-2017-ethical}
Adrian Benton, Glen Coppersmith, and Mark Dredze. 2017.
\newblock \href {https://doi.org/10.18653/v1/W17-1612} {Ethical research
  protocols for social media health research}.
\newblock In \emph{Proceedings of the First {ACL} Workshop on Ethics in Natural
  Language Processing}, pages 94--102, Valencia, Spain. Association for
  Computational Linguistics.

\bibitem[{Br{\aa}dvik(2018)}]{braadvik2018suicide}
Louise Br{\aa}dvik. 2018.
\newblock Suicide risk and mental disorders.

\bibitem[{Brown et~al.(2020)Brown, Mann, Ryder, Subbiah, Kaplan, Dhariwal,
  Neelakantan, Shyam, Sastry, Askell, Agarwal, Herbert-Voss, Krueger, Henighan,
  Child, Ramesh, Ziegler, Wu, Winter, Hesse, Chen, Sigler, Litwin, Gray, Chess,
  Clark, Berner, McCandlish, Radford, Sutskever, and
  Amodei}]{NEURIPS2020_1457c0d6}
Tom Brown, Benjamin Mann, Nick Ryder, Melanie Subbiah, Jared~D Kaplan, Prafulla
  Dhariwal, Arvind Neelakantan, Pranav Shyam, Girish Sastry, Amanda Askell,
  Sandhini Agarwal, Ariel Herbert-Voss, Gretchen Krueger, Tom Henighan, Rewon
  Child, Aditya Ramesh, Daniel Ziegler, Jeffrey Wu, Clemens Winter, Chris
  Hesse, Mark Chen, Eric Sigler, Mateusz Litwin, Scott Gray, Benjamin Chess,
  Jack Clark, Christopher Berner, Sam McCandlish, Alec Radford, Ilya Sutskever,
  and Dario Amodei. 2020.
\newblock \href
  {https://proceedings.neurips.cc/paper/2020/file/1457c0d6bfcb4967418bfb8ac142f64a-Paper.pdf}
  {Language models are few-shot learners}.
\newblock In \emph{Advances in Neural Information Processing Systems},
  volume~33, pages 1877--1901. Curran Associates, Inc.

\bibitem[{Br\r{a}dvik(2018)}]{51109334-209f-4220-b5de-bfad08259c34}
Louise Br\r{a}dvik. 2018.
\newblock \href {https://doi.org/10.3390/ijerph15092028} {Suicide risk and
  mental disorders}.

\bibitem[{Cohan et~al.(2018)Cohan, Desmet, Yates, Soldaini, MacAvaney, and
  Goharian}]{cohan2018smhd}
Arman Cohan, Bart Desmet, Andrew Yates, Luca Soldaini, Sean MacAvaney, and
  Nazli Goharian. 2018.
\newblock Smhd: a large-scale resource for exploring online language usage for
  multiple mental health conditions.
\newblock In \emph{Proceedings of the 27th International Conference on
  Computational Linguistics}, pages 1485--1497.

\bibitem[{Coppersmith et~al.(2014)Coppersmith, Dredze, and
  Harman}]{coppersmith-etal-2014-quantifying}
Glen Coppersmith, Mark Dredze, and Craig Harman. 2014.
\newblock \href {https://doi.org/10.3115/v1/W14-3207} {Quantifying mental
  health signals in {T}witter}.
\newblock In \emph{Proceedings of the Workshop on Computational Linguistics and
  Clinical Psychology: From Linguistic Signal to Clinical Reality}, pages
  51--60, Baltimore, Maryland, USA. Association for Computational Linguistics.

\bibitem[{Coppersmith et~al.(2018)Coppersmith, Leary, Crutchley, and
  Fine}]{Coppersmith2018NaturalLP}
Glen Coppersmith, Ryan Leary, Patrick Crutchley, and A.~Fine. 2018.
\newblock Natural language processing of social media as screening for suicide
  risk.
\newblock \emph{Biomedical Informatics Insights}, 10.

\bibitem[{D'Amour et~al.(2020)D'Amour, Heller, Moldovan, Adlam, Alipanahi,
  Beutel, Chen, Deaton, Eisenstein, Hoffman, Hormozdiari, Houlsby, Hou, Jerfel,
  Karthikesalingam, Lucic, Ma, McLean, Mincu, Mitani, Montanari, Nado,
  Natarajan, Nielson, Osborne, Raman, Ramasamy, Sayres, Schrouff, Seneviratne,
  Sequeira, Suresh, Veitch, Vladymyrov, Wang, Webster, Yadlowsky, Yun, Zhai,
  and Sculley}]{damour2020underspecification}
Alexander D'Amour, Katherine Heller, Dan Moldovan, Ben Adlam, Babak Alipanahi,
  Alex Beutel, Christina Chen, Jonathan Deaton, Jacob Eisenstein, Matthew~D.
  Hoffman, Farhad Hormozdiari, Neil Houlsby, Shaobo Hou, Ghassen Jerfel, Alan
  Karthikesalingam, Mario Lucic, Yian Ma, Cory McLean, Diana Mincu, Akinori
  Mitani, Andrea Montanari, Zachary Nado, Vivek Natarajan, Christopher Nielson,
  Thomas~F. Osborne, Rajiv Raman, Kim Ramasamy, Rory Sayres, Jessica Schrouff,
  Martin Seneviratne, Shannon Sequeira, Harini Suresh, Victor Veitch, Max
  Vladymyrov, Xuezhi Wang, Kellie Webster, Steve Yadlowsky, Taedong Yun,
  Xiaohua Zhai, and D.~Sculley. 2020.
\newblock Underspecification presents challenges for credibility in modern
  machine learning.
\newblock \emph{arXiv preprint arXiv:2011.03395}.

\bibitem[{De~Choudhury et~al.(2013)De~Choudhury, Gamon, Counts, and
  Horvitz}]{de2013predicting}
Munmun De~Choudhury, Michael Gamon, Scott Counts, and Eric Horvitz. 2013.
\newblock Predicting depression via social media.
\newblock \emph{{ICWSM}}, 13:1--10.

\bibitem[{Delahunty et~al.(2019)Delahunty, Johansson, and
  Arcan}]{delahunty-etal-2019-passive}
Fionn Delahunty, Robert Johansson, and Mihael Arcan. 2019.
\newblock \href {https://doi.org/10.18653/v1/W19-3205} {Passive diagnosis
  incorporating the {PHQ}-4 for depression and anxiety}.
\newblock In \emph{Proceedings of the Fourth Social Media Mining for Health
  Applications ({\#}SMM4H) Workshop {\&} Shared Task}, pages 40--46, Florence,
  Italy. Association for Computational Linguistics.

\bibitem[{Devlin et~al.(2019{\natexlab{a}})Devlin, Chang, Lee, and
  Toutanova}]{devlin2019bert}
Jacob Devlin, Ming-Wei Chang, Kenton Lee, and Kristina Toutanova.
  2019{\natexlab{a}}.
\newblock Bert: Pre-training of deep bidirectional transformers for language
  understanding.
\newblock In \emph{Proceedings of the 2019 Conference of the North American
  Chapter of the Association for Computational Linguistics: Human Language
  Technologies, Volume 1 (Long and Short Papers)}, pages 4171--4186.

\bibitem[{Devlin et~al.(2019{\natexlab{b}})Devlin, Chang, Lee, and
  Toutanova}]{devlin-etal-2019-bert}
Jacob Devlin, Ming-Wei Chang, Kenton Lee, and Kristina Toutanova.
  2019{\natexlab{b}}.
\newblock \href {https://doi.org/10.18653/v1/N19-1423} {{BERT}: Pre-training of
  deep bidirectional transformers for language understanding}.
\newblock In \emph{Proceedings of the 2019 Conference of the North {A}merican
  Chapter of the Association for Computational Linguistics: Human Language
  Technologies, Volume 1 (Long and Short Papers)}, pages 4171--4186,
  Minneapolis, Minnesota. Association for Computational Linguistics.

\bibitem[{Ernala et~al.(2019)Ernala, Birnbaum, Candan, Rizvi, Sterling, Kane,
  and De~Choudhury}]{10.1145/3290605.3300364}
Sindhu~Kiranmai Ernala, Michael~L. Birnbaum, Kristin~A. Candan, Asra~F. Rizvi,
  William~A. Sterling, John~M. Kane, and Munmun De~Choudhury. 2019.
\newblock \href {https://doi.org/10.1145/3290605.3300364} {Methodological gaps
  in predicting mental health states from social media: Triangulating
  diagnostic signals}.
\newblock In \emph{Proceedings of the 2019 CHI Conference on Human Factors in
  Computing Systems}, CHI '19, page 1–16, New York, NY, USA. Association for
  Computing Machinery.

\bibitem[{Gardner et~al.(2017)Gardner, Grus, Neumann, Tafjord, Dasigi, Liu,
  Peters, Schmitz, and Zettlemoyer}]{Gardner2017ADS}
Matt Gardner, Joel Grus, Mark Neumann, Oyvind Tafjord, Pradeep Dasigi, Nelson
  H~S Liu, Matthew~E. Peters, Michael Schmitz, and Luke Zettlemoyer. 2017.
\newblock A deep semantic natural language processing platform.

\bibitem[{Geirhos et~al.(2020)Geirhos, Jacobsen, Michaelis, Zemel, Brendel,
  Bethge, and Wichmann}]{Geirhos2020ShortcutLI}
Robert Geirhos, Jorn-Henrik Jacobsen, Claudio Michaelis, Richard Zemel, Wieland
  Brendel, Matthias Bethge, and Felix Wichmann. 2020.
\newblock Shortcut learning in deep neural networks.
\newblock \emph{Nature Machine Intelligence}.

\bibitem[{Go et~al.(2009)Go, Bhayani, and Huang}]{go2009twitter}
Alec Go, Richa Bhayani, and Lei Huang. 2009.
\newblock Twitter sentiment classification using distant supervision.
\newblock \emph{CS224N project report, Stanford}, 1(12):2009.

\bibitem[{Gratch et~al.(2014)Gratch, Artstein, Lucas, Stratou, Scherer,
  Nazarian, Wood, Boberg, DeVault, Marsella et~al.}]{gratch2014distress}
Jonathan Gratch, Ron Artstein, Gale~M Lucas, Giota Stratou, Stefan Scherer,
  Angela Nazarian, Rachel Wood, Jill Boberg, David DeVault, Stacy Marsella,
  et~al. 2014.
\newblock The distress analysis interview corpus of human and computer
  interviews.
\newblock In \emph{LREC}, pages 3123--3128.

\bibitem[{Harrigian et~al.(2020)Harrigian, Aguirre, and
  Dredze}]{harrigian-etal-2020-models}
Keith Harrigian, Carlos Aguirre, and Mark Dredze. 2020.
\newblock \href {https://www.aclweb.org/anthology/2020.findings-emnlp.337} {Do
  models of mental health based on social media data generalize?}
\newblock In \emph{Findings of the Association for Computational Linguistics:
  EMNLP 2020}, pages 3774--3788, Online. Association for Computational
  Linguistics.

\bibitem[{Harrigian et~al.(2021)Harrigian, Aguirre, and
  Dredze}]{harrigian-etal-2021-state}
Keith Harrigian, Carlos Aguirre, and Mark Dredze. 2021.
\newblock \href {https://doi.org/10.18653/v1/2021.clpsych-1.2} {On the state of
  social media data for mental health research}.
\newblock In \emph{Proceedings of the Seventh Workshop on Computational
  Linguistics and Clinical Psychology: Improving Access}, pages 15--24, Online.
  Association for Computational Linguistics.

\bibitem[{Houlsby et~al.()Houlsby, Giurgiu, Jastrzebski, Morrone,
  de~Laroussilhe, Gesmundo, Attariyan, and Gelly}]{houlsbyparameter}
Neil Houlsby, Andrei Giurgiu, Stanis{\l}aw Jastrzebski, Bruna Morrone, Quentin
  de~Laroussilhe, Andrea Gesmundo, Mona Attariyan, and Sylvain Gelly.
\newblock Parameter-efficient transfer learning for nlp.

\bibitem[{Jiang et~al.(2020)Jiang, Levitan, Zomick, and
  Hirschberg}]{jiang2020detection}
Zheng~Ping Jiang, Sarah~Ita Levitan, Jonathan Zomick, and Julia Hirschberg.
  2020.
\newblock Detection of mental health from reddit via deep contextualized
  representations.
\newblock In \emph{Proceedings of the 11th International Workshop on Health
  Text Mining and Information Analysis}, pages 147--156.

\bibitem[{Kroenke et~al.(2001)Kroenke, Spitzer, and Williams}]{kroenke2001phq}
Kurt Kroenke, Robert~L Spitzer, and Janet~BW Williams. 2001.
\newblock The {PHQ-9}: validity of a brief depression severity measure.
\newblock \emph{Journal of general internal medicine}, 16(9):606--613.

\bibitem[{Lee et~al.(2021)Lee, Kummerfeld, An, and
  Mihalcea}]{lee-etal-2021-micromodels-efficient}
Andrew Lee, Jonathan~K. Kummerfeld, Larry An, and Rada Mihalcea. 2021.
\newblock \href {https://doi.org/10.18653/v1/2021.findings-emnlp.360}
  {Micromodels for efficient, explainable, and reusable systems: A case study
  on mental health}.
\newblock In \emph{Findings of the Association for Computational Linguistics:
  EMNLP 2021}, pages 4257--4272, Punta Cana, Dominican Republic. Association
  for Computational Linguistics.

\bibitem[{Losada and Crestani(2016)}]{losada2016test}
David~E Losada and Fabio Crestani. 2016.
\newblock A test collection for research on depression and language use.
\newblock In \emph{International Conference of the Cross-Language Evaluation
  Forum for European Languages}, pages 28--39. Springer.

\bibitem[{Losada et~al.(2019)Losada, Crestani, and
  Parapar}]{losada2019overview}
David~E Losada, Fabio Crestani, and Javier Parapar. 2019.
\newblock Overview of erisk 2019 early risk prediction on the internet.
\newblock In \emph{International Conference of the Cross-Language Evaluation
  Forum for European Languages}, pages 340--357. Springer.

\bibitem[{Matero et~al.(2019)Matero, Idnani, Son, Giorgi, Vu, Zamani,
  Limbachiya, Guntuku, and Schwartz}]{matero2019suicide}
Matthew Matero, Akash Idnani, Youngseo Son, Salvatore Giorgi, Huy Vu, Mohammad
  Zamani, Parth Limbachiya, Sharath~Chandra Guntuku, and H~Andrew Schwartz.
  2019.
\newblock Suicide risk assessment with multi-level dual-context language and
  bert.
\newblock In \emph{Proceedings of the Sixth Workshop on Computational
  Linguistics and Clinical Psychology}, pages 39--44.

\bibitem[{Milne et~al.(2016)Milne, Pink, Hachey, and
  Calvo}]{milne-etal-2016-clpsych}
David~N. Milne, Glen Pink, Ben Hachey, and Rafael~A. Calvo. 2016.
\newblock \href {https://doi.org/10.18653/v1/W16-0312} {{CLP}sych 2016 shared
  task: Triaging content in online peer-support forums}.
\newblock In \emph{Proceedings of the Third Workshop on Computational
  Linguistics and Clinical Psychology}, pages 118--127, San Diego, CA, USA.
  Association for Computational Linguistics.

\bibitem[{Mullenbach et~al.(2018)Mullenbach, Wiegreffe, Duke, Sun, and
  Eisenstein}]{mullenbach-etal-2018-explainable}
James Mullenbach, Sarah Wiegreffe, Jon Duke, Jimeng Sun, and Jacob Eisenstein.
  2018.
\newblock \href {https://doi.org/10.18653/v1/N18-1100} {Explainable prediction
  of medical codes from clinical text}.
\newblock In \emph{Proceedings of the 2018 Conference of the North {A}merican
  Chapter of the Association for Computational Linguistics: Human Language
  Technologies, Volume 1 (Long Papers)}, pages 1101--1111, New Orleans,
  Louisiana. Association for Computational Linguistics.

\bibitem[{Peng et~al.(2019)Peng, Yan, and Lu}]{peng2019transfer}
Yifan Peng, Shankai Yan, and Zhiyong Lu. 2019.
\newblock Transfer learning in biomedical natural language processing: An
  evaluation of bert and elmo on ten benchmarking datasets.
\newblock \emph{BioNLP 2019}, page~58.

\bibitem[{Ramirez-Esparza et~al.(2008)Ramirez-Esparza, Chung, Kacewicz, and
  Pennebaker}]{ramirez2008psychology}
Nairan Ramirez-Esparza, Cindy~K Chung, Ewa Kacewicz, and James~W Pennebaker.
  2008.
\newblock The psychology of word use in depression forums in english and in
  spanish: Texting two text analytic approaches.
\newblock In \emph{ICWSM}.

\bibitem[{Rietzler et~al.(2020)Rietzler, Stabinger, Opitz, and
  Engl}]{rietzler2020adapt}
Alexander Rietzler, Sebastian Stabinger, Paul Opitz, and Stefan Engl. 2020.
\newblock Adapt or get left behind: Domain adaptation through bert language
  model finetuning for aspect-target sentiment classification.
\newblock In \emph{Proceedings of the 12th Language Resources and Evaluation
  Conference}, pages 4933--4941.

\bibitem[{Rinaldi et~al.(2020)Rinaldi, Tree, and
  Chaturvedi}]{rinaldi2020predicting}
Alex Rinaldi, Jean E~Fox Tree, and Snigdha Chaturvedi. 2020.
\newblock Predicting depression in screening interviews from latent
  categorization of interview prompts.
\newblock In \emph{Proceedings of the 58th Annual Meeting of the Association
  for Computational Linguistics}, pages 7--18.

\bibitem[{Shing et~al.(2018)Shing, Nair, Zirikly, Friedenberg, Daum{\'e}~III,
  and Resnik}]{shing2018expert}
Han-Chin Shing, Suraj Nair, Ayah Zirikly, Meir Friedenberg, Hal Daum{\'e}~III,
  and Philip Resnik. 2018.
\newblock \href {https://doi.org/10.18653/v1/W18-0603} {Expert, crowdsourced,
  and machine assessment of suicide risk via online postings}.
\newblock In \emph{Proceedings of the Fifth Workshop on Computational
  Linguistics and Clinical Psychology: From Keyboard to Clinic}, pages 25--36,
  New Orleans, LA. Association for Computational Linguistics.

\bibitem[{Shing et~al.(2020)Shing, Resnik, and Oard}]{shing2020prioritization}
Han-Chin Shing, Philip Resnik, and Douglas~W Oard. 2020.
\newblock A prioritization model for suicidality risk assessment.
\newblock In \emph{Proceedings of the 58th Annual Meeting of the Association
  for Computational Linguistics}, pages 8124--8137.

\bibitem[{Tausczik and Pennebaker(2010)}]{tausczik2010psychological}
Yla~R Tausczik and James~W Pennebaker. 2010.
\newblock The psychological meaning of words: Liwc and computerized text
  analysis methods.
\newblock \emph{Journal of language and social psychology}, 29(1):24--54.

\bibitem[{Wolohan et~al.(2018)Wolohan, Hiraga, Mukherjee, Sayyed, and
  Millard}]{Wolohan2018DetectingLT}
Jt~Wolohan, Misato Hiraga, Atreyee Mukherjee, Z.~Sayyed, and Matthew Millard.
  2018.
\newblock Detecting linguistic traces of depression in topic-restricted text:
  Attending to self-stigmatized depression with nlp.

\bibitem[{Yadav et~al.(2020)Yadav, Chauhan, Sain, Thirunarayan, Sheth, and
  Schumm}]{yadav-etal-2020-identifying}
Shweta Yadav, Jainish Chauhan, Joy~Prakash Sain, Krishnaprasad Thirunarayan,
  Amit Sheth, and Jeremiah Schumm. 2020.
\newblock \href {https://doi.org/10.18653/v1/2020.coling-main.61} {Identifying
  depressive symptoms from tweets: Figurative language enabled multitask
  learning framework}.
\newblock In \emph{Proceedings of the 28th International Conference on
  Computational Linguistics}, pages 696--709, Barcelona, Spain (Online).
  International Committee on Computational Linguistics.

\bibitem[{Yates et~al.(2017)Yates, Cohan, and
  Goharian}]{yates-etal-2017-depression}
Andrew Yates, Arman Cohan, and Nazli Goharian. 2017.
\newblock \href {https://doi.org/10.18653/v1/D17-1322} {Depression and
  self-harm risk assessment in online forums}.
\newblock In \emph{Proceedings of the 2017 Conference on Empirical Methods in
  Natural Language Processing}, pages 2968--2978, Copenhagen, Denmark.
  Association for Computational Linguistics.

\bibitem[{Zirikly et~al.(2019)Zirikly, Resnik, Uzuner, and
  Hollingshead}]{zirikly2019clpsych}
Ayah Zirikly, Philip Resnik, Ozlem Uzuner, and Kristy Hollingshead. 2019.
\newblock {CLPsych} 2019 shared task: Predicting the degree of suicide risk in
  {Reddit} posts.
\newblock In \emph{Proceedings of the Sixth Workshop on Computational
  Linguistics and Clinical Psychology}, pages 24--33.

\end{thebibliography}
\bibliographystyle{acl_natbib}
\newpage
\appendix

\section{Details of experimental setup}
We designed various experiments to validate and analyze two main components: the questionnaire and depression models. The hyper-parameters of those models were set as follows:

\textbf{Questionnaire model}: In classifier-constrained methods, we trained a CNN for each of the nine symptom classifiers. We used filters of sizes $[2, 3, 4, 5, 6]$, and one filter for each size. Consequently, the max-pooling produces a vector of size $5$, which is then fed into the final linear layer for prediction. We apply $L_2$ regularization specifically to the CNN kernels. The $L_2$ weights were fine-tuned with three options [$0.1$, $0.01$, $0.001$]. 

\textbf{Depression model}: We experimented with $6$ variations described in Table \ref{tab:exp_variations}. 
% These correspond to Sections \ref{pbc}, \ref{qbm}, and \ref{unconstrained_md}, respectively.
All of these variations, except for the first one, use a CNN classifier on top of different inputs ranging from BERT embedding to pattern scores. The CNN used here has ($filter\_size = [2,3,4,5,6]$, $num\_filter = 50$), and $(k=5)$ for k-max pooling. The threshold for the first variation was tuned on from $1$ to $10$. 

In all experiments with pre-trained BERT, we used the BERT-base version \cite{devlin2019bert}. We do not fine-tune BERT's parameters since in a pilot study, we found that fine-tuning BERT does not improve the generalization. All models were optimized using a cross-entropy loss with class weights of $0.1$ and $0.9$ for the control and depressed classes, respectively, a 1cycle learning rate scheduler (with a maximum of $0.01$), batches of size 64, and early stopping after five epochs. When calculating F1, we set the decision threshold to $0.5$, because it cannot be safely tuned in dataset-transfer experiments.

\section{Manually-labeled questionnaire dataset} \label{sec:manual_dataset}
To evaluate the performance of our weakly-supervised symptom classifiers, we prepared 900 examples manually labeled by three annotators. 
The labeled samples were randomly selected posts containing carefully selected keywords (e.g., keywords that are close to positive patterns - ``\textit{sleep}`` in ``\textit{can't sleep}``) to avoid including too many easy true negatives. The labeling process involved three annotators. The first annotator labeled all 100 instances, and the second annotator re-labeled 50 of them. If the agreement on twice-labeled examples was weak ($\kappa < 0.60$), the second annotator would continue to annotate the remaining 50 examples. The third annotator adjudicated label disagreements between the first two annotators.

\section{Details of questionnaire dataset construction} \label{data-construction}
This section describes the data creation process for the questionnaire models, which consists of 9 symptom classifiers corresponding to 9 questions in the PHQ9 instrument (e.g., \textit{``trouble falling or staying asleep?''}). PHQ9 questions ask how often the patient experienced each symptom; we adapt this approach to our domain by classifying whether a given post contains a symptom.
Given the lack of training data for this task, we collected regular expression patterns and used these patterns together with heuristics to construct training data (positive class and negative class) for each symptom. 
% \begin{figure*}[h!]
%     \centering
%     \includegraphics[width=0.6\linewidth]{figures/methods.pdf}
%     \caption{Weakly-supervised questionnaire model}
%     \label{fig:symptom_models}
%     \vspace{-2mm} 
% \end{figure*}
% \subsection{Pattern-based generated data}
\subsection{Positive class}
For each question in the PHQ9 diagonstic instrument, we prepared a set of positive patterns that each indicates the presence of a symptom described by the question. For example, the patterns ``\textit{don'?t feel like doing anything}'' and ``\textit{can'?t fall back to sleep}'' describe the anhedonia and sleep symptoms, respectively. The number of patterns for each symptom is shown in Table \ref{tab:pattern_stat}. Each pattern set is then matched against a collection of posts crawled from 127 mental-health subreddits\footnote{\url{https://files.pushshift.io/reddit/}} and also against posts from the SMHD dataset \cite{cohan2018smhd}, which was constructed from Reddit but excludes mental health subreddits. 
The purpose of using these two raw datasets is to increase the diversity and minimize the bias of the data.
In the labeling step, if a post contains a match with any positive pattern of a question, we select that post as a positive training sample for the corresponding symptom question.
% \begin{table}[t]
%     \centering
%     \small
%     \begin{tabular}{ll}
%     \toprule
%     \textbf{Question} & \textbf{Patterns}\\
%     \midrule
%     \multirow{3}*{\textbf{Anhedonia}} & (couldn't|could not) enjoy anything\\
%     & can't find.\{0,15\} motivation \\
%     & don'?t feel like doing anything \\
%     \hline 
%     \multirow{3}*{\textbf{Sleep}} & can'?t fall back to sleep \\
%     & haven'?t slept for( many)* days \\
%     & constantly waking up\\
%     \bottomrule 
%     \end{tabular}
%     \caption{Examples of positive patterns}
%     \label{tab:patern_sample}
% \end{table}
\begin{table}[t]
    \small
    \begin{tabular}{lrlr}
    \toprule
    \textbf{Symptom} & \textbf{\# patterns} & \textbf{Symptom} & \textbf{\# patterns} \\
    \cmidrule(lr){1-2} \cmidrule(lr){3-4}
    Anhedonia & 116 & Psychomotor & 108 \\
    Concentration & 70 & Self-esteem & 102 \\
    Eating & 145 & Self-harm & 126\\
    Fatigue & 73 & Sleep & 118\\ 
    Mood & 110\\ 
    \bottomrule 
    \end{tabular}   
    \caption{Number of patterns for each question}
    \label{tab:pattern_stat}
    \vspace{-5mm}
\end{table}

While pattern matching is fast and transparent, it is inflexible and may produce many false positives (FP).
Below, we introduce the four most popular FP cases discovered in our analysis and the techniques we employed to mitigate them when constructing weakly-labeled training data.

\paragraphHd{Positive sentiment.}
Some posts contain positive patterns but do not show depressive signal; for example, ``\textit{Friends and I stayed up all night playing a game}'' contains the positive pattern ``\textit{stayed up all night}'', but it shows excitement about the game rather than a sleep issue. As a solution, we removed all posts containing positive sentiment with the help of Allen NLP's sentiment model~\cite{Gardner2017ADS}.  

\paragraphHdNospace{Conditional clause.}
Sometimes users hypothesize about their health conditions, such as ``\textit{If I lost my appetite for days at a time, that...  wouldn't be sustainable for me.}''.
We remove these posts by using regular expressions to identify popular conditional clause formats.

\paragraphHdNospace{Third-person pronouns.} 
Users may attribute a condition to someone else, such as in ``\textit{he is easily distracted.}''
We identified posts of this kind by checking whether the closest pronoun to the positive pattern is third-person or first-person, and removing posts in the former category.
% This includes references like ``\textit{my son}''.
%similar to the above example by checking whether the pronoun (``\textit{My son}'') closest to the positive pattern (``\textit{easily distracted}'') is third-person or first-person pronoun. 

\paragraphHdNospace{Negation.} Positive patterns may be negated, such as in ``\textit{haven't had a suicidal thought in ages.}''. To handle this situation, we removed posts containing positive patterns preceded by a negation word, such as (``\textit{not}'', ``\textit{never}'', `\textit{rarely}'', etc.). 

% While these heuristics removed many false positives, there are still harder cases that require semantic understanding of the text.
% For example, ``\textit{it's \underline{difficult to focus} on my breathing for very long}'' and  ``\textit{that behavior makes me \underline{lose my appetite}}''. We leave these difficult cases for future work.

\subsection{Negative class}
Identifying hard negative samples is crucial for the quality of the trained classifiers.
Models trained on negative examples that are too easy might be prone to over-fitting or perform only keyword matching. Therefore, we propose five heuristics for identifying and synthesizing negative examples.

\paragraphHd{Keywords.}
We collect negative posts that contain some keywords, such as \textit{"sleep"}, but do not contain a positive pattern (``\textit{can't sleep}'').
This hinders models to perform simple keyword matching.
%The resulted posts are labeled as negative. Doing this provides harder negative examples, which in turn would hinder the trained model from doing keyword matching. 
 
\paragraphHdNospace{Pronouns.}
We replace the first-person pronouns appearing in posts from the positive class with third-person pronouns or proper nouns, such as replacing ``\textit{I}'' with ``\textit{she}.''
%This method could help further removing the FP case about third-person pronoun described in Section \ref{positive_class_data}. 

\paragraphHdNospace{Other symptoms.}
We use randomly selected posts labeled positive for other questions (symptoms) as negative examples for the given question.
We ensure the selected posts do not contain positive patterns for the given question.
%positive class from other questions could also be used as negative examples of the considered question. We also have to make sure that the selected posts do not contain positive patterns of the current question. 

\paragraphHdNospace{Negation.}
For each positive pattern defined in the previous section, we created a corresponding negated one, such as negating ``\textit{have sleep apnea}'' to ``\textit{never have sleep apnea}''.
%Given this mapping, we loop through all sentences in posts of positive class, select those containing a positive pattern in the mapping, negate and use them as negative samples.
Only matched sentences were used in this method, because using the whole post with only some sentences being negated could lead to contextual inconsistencies.
%, which in turn harming the trained classifier.  

\paragraphHdNospace{Positive sentiment.}
We use training instances labeled neutral or positive in a sentiment dataset as negative examples.
In particular, we used the Sentiment140 corpus, which contains 1.6 millions tweets \cite{go2009twitter}.

\end{document}